\documentclass[11pt,a4paper]{article}
\usepackage[hyperref]{ranlp2021}
\usepackage{times}
\usepackage{multirow}
\usepackage{latexsym}

\usepackage{url}
\usepackage[T1]{fontenc}

\usepackage{url}
\usepackage{multirow}
\usepackage{booktabs}
\usepackage{colortbl}
\usepackage{graphicx}

\usepackage{graphicx} 
\usepackage{color}

\usepackage{amsmath}
\usepackage{amssymb}
\usepackage{algorithm}
\usepackage{algorithmic}
\usepackage{marvosym}

\usepackage[belowskip=5pt,aboveskip=0pt]{caption}

\usepackage[T1]{fontenc}

\usepackage[utf8]{inputenc}
\usepackage{ulem}

\usepackage{microtype}

\aclfinalcopy 

\title{AutoChart: A Dataset for Chart-to-Text Generation Task}
\newcommand*{\affaddr}[1]{#1} 
\newcommand*{\affmark}[1][*]{\textsuperscript{#1}}
\newcommand*{\email}[1]{\texttt{#1}}

\author{%
Jiawen Zhu\affmark[1], Jinye Ran\affmark[3], Roy Ka-wei Lee\affmark[1], Kenny Choo\affmark[1], and Zhi Li\affmark[2]\\
\affaddr{\affmark[1]Singapore University of Technology and Design}\\
\affaddr{\affmark[2]University of Saskatchewan}\\
\affaddr{\affmark[3]China Merchants Bank}\\
\email{\{jiawen\_zhu,roy\_lee,kenny\_choo\}@sutd.edu.sg}\\
\email{z.li@usask.ca},
\email{rjy777@163.com}\\
}

\begin{document}
\maketitle
\begin{abstract}
The analytical description of charts is an exciting and important research area with many applications in academia and industry. Yet, this challenging task has received limited attention from the computational linguistics research community. 
This paper proposes \textsf{AutoChart}, a large dataset for the analytical description of charts, which aims to encourage more research into this important area. 
Specifically, we offer a novel framework that generates the charts and their analytical description automatically. We conducted extensive human and machine evaluations on the generated charts and descriptions and demonstrate that the generated texts are informative, coherent, and relevant to the corresponding charts.
\end{abstract}
\section{Introduction}
\label{sec:introduction}
Natural language generation (NLG) is one of the core research areas in artificial intelligence \cite{gatt2018survey}. Recent NLG studies have explored data-to-text generation, where exciting applications such as automated news reporting \cite{leppanen2017data} were developed to generate text from non-linguistic data automatically. In this paper, we explore the \textbf{chart-to-text} generation problem, where analytical textual descriptions are automatically generated for a given graphical chart. 

\begin{figure}[t]
\centering
\begin{tabular}{p{18em}}
\multicolumn{1}{c}{\includegraphics[scale=0.18]{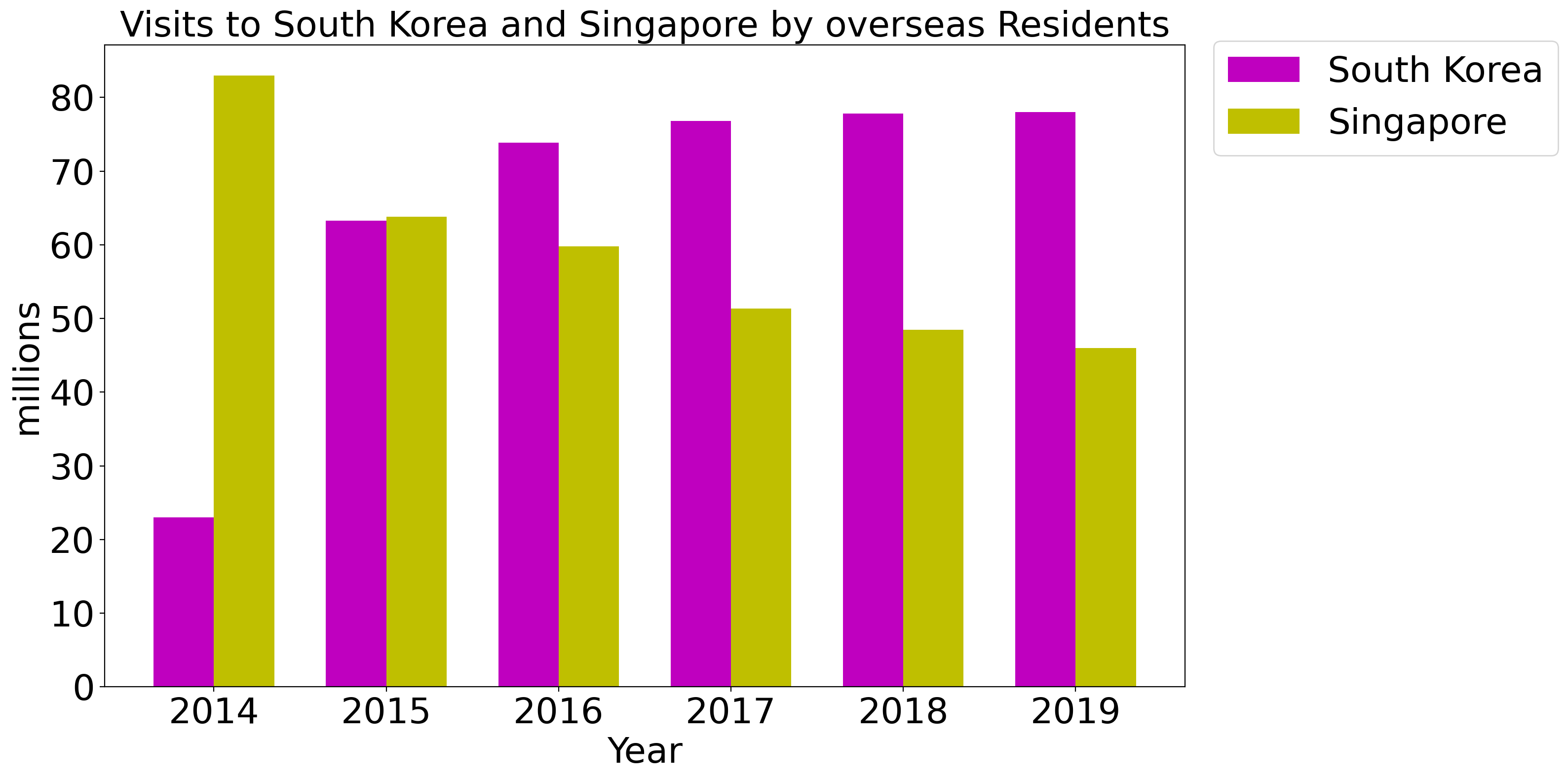}}\\ 
\textit{\cellcolor[HTML]{EFEFEF} \small This bar graph shows the number of visits to South Korea and Singapore by overseas residents, respectively, from 2014 to 2019. In 2014, there was a huge gap in the number of visits to these two countries. The number of visits to South Korea is about 20 million,  whereas the number of visits to Singapore is over 80 million. There is a continuous decrease in the number of visits to Singapore, with the largest decrease in 2015 to about 60 million. In 2019, the number of visits becomes about 45 million. By contrast, the number of visits to South Korea has been on the rise since 2014 but seems to have plateaued in 2017.}\\
\end{tabular}
\caption{Example of IELTS AWT1.}
\label{fig:example}
\end{figure}

Chart-to-text generation has many exciting academic and commercial applications. For instance, preliminary analyses can be generated on charts to aid users in authoring analytical documents. On the accessibility front, automatically generated chart analyses can also support accessibility since text descriptions can be fed into speech-to-text modules and help visually impaired individuals to understand charts. Chart-to-text generation could also be applied to aid academic writing. Text descriptions of visual elements such as diagrams, charts, and graphs, are among the core academic assignments in linguistics \cite{molle2008multimodal}. For example, the IELTS Academic Writing Task 1 (AWT1) is an assessment task that elicits written responses on a visual-verbal relationship. The AWT1 requires test takers to ``\textit{describe, summarise, or explain the information in a graph, table, chart, or diagram.}'' Figure~\ref{fig:example} shows an example of the AWT1. Chart-to-text generation offers the potential to generate large-scale chart analytical description learning examples for students attempting AWT1.

Despite the many benefits and applications of chart-to-text generation, this NLG task has received limited attention from computational linguistics and NLG researchers. Among the key factors that hinder the development of this research area is the lack of a large chart description dataset that may facilitate chart description studies. Intuitively, one possible solution is to collect and manually annotate a chart description. For instance, we will first need to obtain a large dataset of charts and subsequently engage human annotators to write the summary and explanations for these charts. However, such a data collection process is time-consuming and expensive. Another approach is to perform large-scale data-crawling to retrieve charts and corresponding human-written summaries from the Internet. However, it is challenging to ensure that text summaries correctly describe the chart and have provided adequate details to aid readers in understanding the chart as the charts are retrieved from multiple sources. For instance, in a recent study, \cite{obeid2020chart} had performed a large-scale data collection of charts and corresponding text descriptions. However, the descriptions of the chart in the dataset contained background knowledge beyond the data illustrated in the chart.


In this paper, we aim to address chart analysis data scarcity and quality problems by proposing a novel framework that generates charts and their corresponding high-quality descriptions \textit{automatically}. The \textsf{AutoChart}\footnote{Code: https://gitlab.com/bottle\_shop/snlg/chart/autochart} dataset generated by our proposed framework will pioneer new computational linguistic and NLG research area on chart descriptions. For instance, the availability of a large-scale chart description dataset encourages the creation of supervised machine learning and NLP models to interpret the charts and generate relevant text descriptions automatically. 

We summarize our contributions as follows: 
\begin{itemize}
    \item We propose a novel framework to generate charts and their corresponding analytical descriptions automatically.
    \item Using our novel framework, we constructed \textsf{AutoChart}, which is a large-scale chart description dataset, and make this openly available to encourage future research.
    \item We conducted extensive human and machine evaluation on the generated charts and descriptions and demonstrate that the generated text is informative, coherent, and relevant to the corresponding charts.
\end{itemize}

\section{Related Work}
\label{sec:related}
There are very few data-to-text works that investigate chart recognition and understanding. 
Many of these existing works focused on extracting data from the various types of visual charts using deep learning computer vision and object recognition techniques \cite{cliche2017scatteract,balaji2018chart,liu2019data,ma2018scatternet,dai2018chart,battle2018beagle,chai2020crowdchart}. For instance, Balaji et al.~\shortcite{balaji2018chart} proposed an automated system that extracted data points from bar and pie charts to create textual descriptions. However, the generated textual descriptions listed data values extracted from the figures in a static format without any analytical discussion about the charts' overall trends or summary. Another line of work have also proposed \textit{table-to-text} models~\cite{Iso2019Learning,DBLP:journals/corr/abs-1809-00582}, which aims to generate long and good-quality description from structured data formatted in a table. Nevertheless, these \textit{table-to-text} models are designed for specific domains and structured data, and it is challenging to adopt these methods in the chart-to-text task.

Another related sub-domain of work is the visual-based question and answer (Q\&A) task. Kahou et al. \shortcite{kahou2017figureqa} introduced the FigureQA corpus, which consists of over one million question-answer pairs grounded in over 100,000 visual charts. Methani et al.~\shortcite{methani2020plotqa} extended the work in \cite{kahou2017figureqa} and proposed the PlotQA corpus, which is a larger dataset with 28.9 million question-answer pairs over 224,377 charts from real-world sources and questions based on crowd-sourced question templates. While large datasets have been collected for the visual-based Q\&A task, these datasets are not applicable to generate analytical chart descriptions as the question-answer pairs are often short and data-specific without any in-depth analysis on the charts. 

Closer to our work, Obeid and Hoque~\shortcite{obeid2020chart} introduced a new large-scale corpus on chart summarization and proposed a transformer-based chart-to-text model. However, the descriptions of the chart in the dataset contained background knowledge beyond the data illustrated in the chart. These '' noises'' from the beyond-chart-data information may affect the learning of text generation models. Another prominent data source, \textit{Statista}, has high-quality charts, but corresponding summaries may not be descriptive of the chart. 


\begin{figure*}[!h]
    \centering
		\includegraphics[scale=0.6]{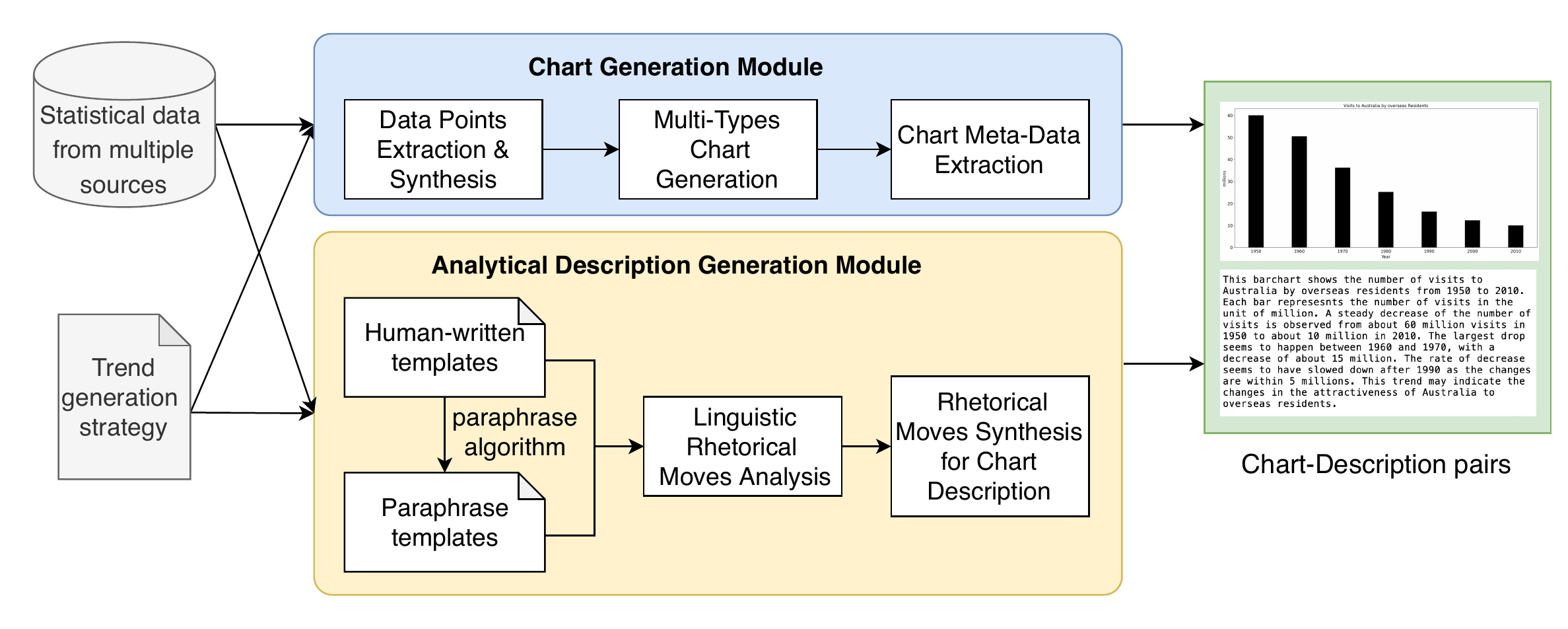}
	\caption{Overview of the AutoChart dataset construction process.}
	\label{fig:framework}
\end{figure*}

Our study addresses the limitations of existing chart-to-text datasets. It extends the existing works on chart recognition and data extraction by offering a novel framework to generate charts and their corresponding analytical descriptions automatically. To this end, we construct and contribute \textsf{AutoChart}, a large-scale chart analytical description dataset.

\section{AutoChart Dataset Construction}
\label{sec:construction}

The goal of this study is to construct a dataset of charts with their corresponding analytical descriptions automatically. To this end, we propose a novel framework to construct the \textsf{AutoChart} dataset and illustrate its construction in Figure \ref{fig:framework}. We begin by collecting statistical data from multiple sources over the web and create the trend generation strategy. The goal of the strategy is to ensure that the generated charts exhibit some form of temporal trends, which ultimately encourages writers to identify these trends analytically. The proposed framework contains two main generation modules: \textit{chart generation} and \textit{analytical description generation}. 

The statistical data and trend generation strategy guide the automatic generation of charts and their meta-information in the chart generation module. Specifically, we generate four types of charts: \textit{scatter plots, line charts, vertical and horizontal bar charts}. 

In the analytical description generation module, linguistic researchers are first recruited to write the analytical descriptions for a few charts. The human-written descriptions are used as templates for the automatic generation of analytical descriptions. As it is labor-intensive to draft human-written descriptions templates, we expand the number of templates by leveraging open-source algorithms to paraphrase the human-written descriptions. Subsequently, we analyze the linguistic rhetorical moves of the human-written and paraphrased templates. The rhetorical move analysis enables us to categorize the rhetorical function types of sentences presented in the analytical description templates. 

Finally, the template sentences annotated with rhetorical moves are strategically sampled and adapted to chart data to generate the analytical description for a given chart. 

\subsection{Statistical Data Collection}
To generate the charts, we first collected statistical data from multiple sources on the web, such as the World Bank Open Data and Nutritional Analysis Data. We crawled data from these sources to extract different variables whose relations could then be plotted (for example, a country's labor force over time, etc.). There are a total of 346 unique indicator variables ($\mathrm{CO_2}$ emission, GDP growth, total population, etc.) with 76 unique entities (cities, states, countries, etc.). The data ranges from 1950 to 2016, though not all indicator variables have data items for all years. The data contains positive integers, floating-point values, and percentages. These values range from 0 to 3.50e+15.



\subsection{Trend Generation Strategy}

Besides plotting the actual collected statistical data, we also aim to generate charts with specific trends. This encourages writers or machine learning algorithms to generate descriptions that analyze the patterns observed in the charts. To this end, we formulate a trend generation strategy, where data perturbation is applied to generate various types of trends. Specifically, we applied the following data perturbation:

\begin{equation}
    Y = S_0e^{(\mu - \frac{\sigma^2}{2})x+\sigma W}
\label{eq:trends}
\end{equation}

\begin{figure}[t]
    \centering
		\includegraphics[width=0.48\textwidth]{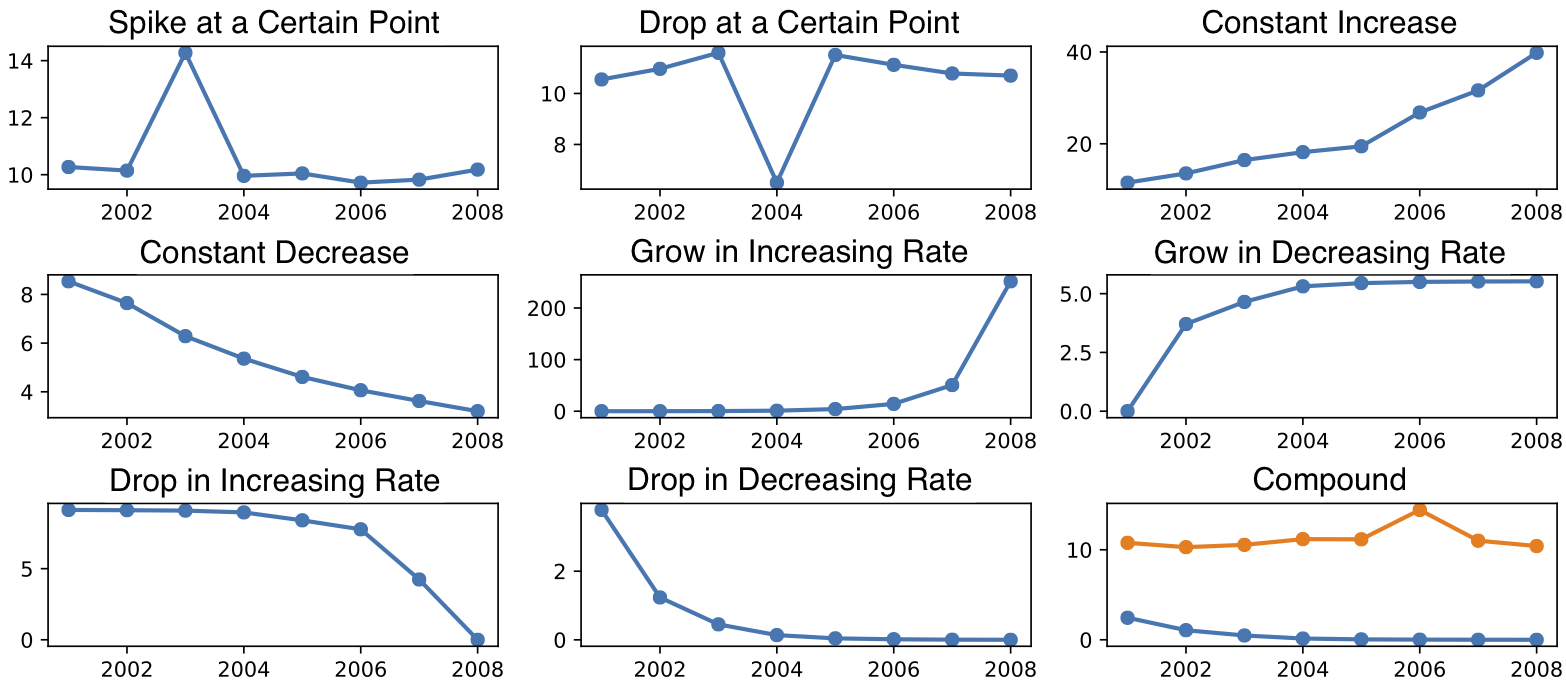}
	\caption{Types of trends generated in AutoChart.}
	\label{fig:trends}
\end{figure}

Here $W$ denotes Brownian motion \cite{brownianmotion} that allows some degree of randomness in the trend generation, $S_0$ denotes the given initial value, $\sigma$ denotes the weight of Brownian motion, that is, the volatility rate of the data. $\mu - \frac{\sigma^2}{2}$ is the drift factor of Brownian motion, which indicates the trend of the data. When it is a positive number, the data is on an increasing trend, and when it is a negative number, it is on a decreasing trend. However, a random fluctuation is generated when it is 0. In total, we apply Equation \ref{eq:trends} to generate charts with eight different types of trends. This is achieved by incorporating various parameters mentioned above and performing symmetry and rotation operations on the data. Figure \ref{fig:trends} shows an example of line charts generated in various trends.

\subsection{Chart Generation}


We generate four types of charts in our \textsf{AutoChart} dataset: \textit{scatter plots, line chart, vertical, and horizontal bar charts}. These types of charts are commonly encountered in academic journals, research papers, textbooks, etc. 

Python library Matplotlib \cite{matplotlib} is used to generate the charts. To encourage diversity in our chart generation, we developed a script to select parameters randomly to add variation to our charts. Specifically, we randomly select markers from 10 unique shapes for each scatter-plot. We also randomly choose the color of the markers in scatter-plots, lines in line charts, and bars in bar charts from a set of 20 colors. The thickness of the bars and line style of lines are also randomly configured. Note that although we fix the size of the entire visual canvas, the size of legends and y-axis values is different for each chart, resulting in random image sizes. The number of discrete elements of x-axis varies from 2 to 8 and the number of entries in legend box varies from 1 to 2. By using different combinations of indicator variables, entities (years, countries, etc.), and parameters, we created a total of 10,232 charts.



Our script preserves the meta-information of the generated charts in JSON files to enable the development of supervised modules for various sub-tasks. Specifically, the meta-information contains bounding box annotations for the legend boxes, legend names and markers, axes labels, axes ticks, data coordinates, plot title, and image index. The meta-information will be used in the analytical description generation module to generate the charts' corresponding descriptions. Furthermore, the meta-data could also be used in evaluating the correctness of future chart understanding models. 

\subsection{Analytical Description Generation}
The creation of analytical descriptions for the generated charts is a challenging task. Firstly, as we have created a large number of charts, it is labor-intensive and time-consuming to draft the analytical descriptions for all the charts manually. Therefore, we would need an automated approach to generate the charts' analytical description. Secondly, the automated solution would need to generate analytical descriptions that are informative, coherent, and relevant to chart context. We propose a template-based approach with linguistics analysis to guide the generation of charts' analytical descriptions to overcome these challenges.


\subsubsection{Templates Generation}
We recruited three linguistics researchers to write the descriptions of a small subset of the generated charts to create the analytical description templates. The subset of generated charts is evenly sampled from the various types of trends. The linguistics researchers are instructed to assume the same setting as IELTS AWT1 when writing the analytical descriptions of sampled charts. In total, the linguistics researchers wrote analytical descriptions for 150 charts. 

As writing the analytical descriptions templates is a labor-intensive and time-consuming task, we used Quillbot\footnote{https://quillbot.com/}, an online paraphrase API, to paraphrase the sentences in the human-written templates. The paraphrase sentences significantly expanded our analytical description templates. In total, we extracted 213 human-written chart sentences, 661 paraphrased sentences as templates. Finally, both human-written and paraphrased sentences will be used to generate other generated charts' analytical descriptions automatically. 

\subsubsection{Rhetorical Move Analysis}
A naive and straightforward way to generate the charts' analytical descriptions is to randomly sample the sentences from our templates and apply the charts' meta-data to produce the relevant analytical descriptions. However, such an approach neglects the rhetorical moves in analytical descriptions, which are important linguistics elements in building analytical arguments \cite{swales2004research}. Inspired by the idea of $moves$ from Swales' framework of genre analysis, we explored a rhetorical moves framework in analytical description templates. Specifically, we manually annotate each sentence in the template and group them in one of the following five rhetorical moves:

\begin{itemize}
\item[(1)] \textbf{Move 1} [Obligatory]: Overview of the chart. This move is used to explain what the chart is about, the chart's content, etc. For example, ``\textit{The chart shows the amount of fast-food consumed in the UK between 1970 and 1990.}''.

\item[(2)] \textbf{Move 2} [Optional]: Description of the chart itself. This move mainly focuses on the configuration of or elements in the chart. For example, ``\textit{All the sampled countries are from Europe: Finland, France, Georgia, Germany, Greece, and Hungary}.''. 

\item[(3)] \textbf{Move 3} [Obligatory]: Interpretation of the chart information. This part mainly explains the changing trend and simple observation of chart information, etc. For example, ``\textit{The amount of fish and chips eaten declined slightly}''.  Nevertheless, it is inadequate to simply describe the trends. Thus, we will add a supplementary \textbf{Move 3.1} to report the numeric information from the chart. For example, ``\textit{In 1970, the consumption was about 300g per week. This fell to 220g per week in 1990.}''. We noted that \textbf{Move 3.1} could be further divided into descriptions of individual data points and comparisons for trends.

\item[(4)] \textbf{Move 4} [Optional]: Evaluative comments on specific value(s) or comparisons. For example, ``\textit{The retired and unemployed people enjoyed about 78 to 82 hours per week which is longer than people from other employment statuses.}''.

\item[(5)] \textbf{Move 5} [Obligatory]: Conclusions, summaries or implications based on the chart. For example, ``\textit{In conclusion, although there was a big increase in the consumption of pizza, sales of fish and chips decreased.}''.
\end{itemize}


In particular, for sentences annotated as Move 3 or 4, we further categorize the sentences into the types of charts that they are describing: 

\begin{itemize}
\item For temporal charts where the x-axis represents time, the sentences focus on the trend of the data and the comparison of different time points. Move 3 and 4 sentences that describe trends are grouped into the eight categories showed in Figure~\ref{fig:trends}. For temporal charts without apparent trends, the sentences will mainly focus on the comparison between data and some special points.

\item For categorical charts where the x-axis represents entities, such as cities, food, etc., the Move 3 and 4 sentences will only focus on comparing different categories and describing some special points.
\end{itemize}


\subsubsection{Rhetorical Moves Synthesis for Chart Description}
After analyzing and annotating the rhetorical moves of sentences in the human-written and paraphrase templates, we leverage the templates' sentences and utilize charts' meta-information to generate the charts' analytical descriptions. To this end, we designed a script that takes in a generated chart as input and performs the following steps:

\begin{table*}[t]
\small
\centering
\begin{tabular}{cc|c|c|c|c|c}
\hline
\multicolumn{1}{l}{}                                     &                                      & \multicolumn{1}{c|}{\multirow{1}{*}{ \textbf{Line}}} & \multicolumn{2}{c|}{\textbf{Bar}}       & \multicolumn{1}{c|}{\multirow{1}{*}{\textbf{Scatter}}} & \multicolumn{1}{c}{\multirow{1}{*}{\textbf{\#Description}}} \\ \cline{4-5}
\multicolumn{1}{l}{}                                     &                                      & \multicolumn{1}{c|}{}   &  \textbf{Horizontal} & \textbf{Vertical} & \multicolumn{1}{c|}{}                                  & \multicolumn{1}{c}{}                                        \\ \hline\hline
\multicolumn{1}{c|}{\multirow{1}{*}{\textbf{Temporal}}} & \multicolumn{1}{c|}{\textbf{Trend}}  & 880                                                 & 480                 & 880               & 880                                                    & 6,805                                                        \\ \cline{2-7} 
\multicolumn{1}{c|}{}                                   & \multicolumn{1}{c|}{\textbf{Random}} & 1,049                                               & 676                 & 1,049             & 1,049                                                  & 9,174                                                        \\ \hline
\multicolumn{2}{c|}{\textbf{Categorical}}                                                      & 951                                                 & 436                 & 951               & 951                                                    & 7,564                                                        \\ \hline
\multicolumn{2}{c|}{\textbf{Total}}                                                            & 2,880                                      & 1,592      & 2,880    & 2,880                                         & 23,543                                              \\ 
\hline
\end{tabular}
\caption{Summary Statistics of \textsf{AutoChart} Dataset.}
\label{tbl:pltbreakdown}
\end{table*}

\begin{enumerate}
    \item We first extract the generated chart's data values and meta-information from its corresponding JSON file. Specifically, we extract the title, x-axis, and y-axis labels, numeric information, the data trend, etc.
    \item Depending on the type of charts (i.e., temporal or categorical), and the trend(s) in the chart, we sample the sentences from the templates such that the sentences of various rhetorical moves are selected to build a coherent analytical description. Furthermore, to encourage diversity in the generated analytical description, we randomly set the number of rhetorical move sentences to generate. The conditional sampling of template sentences by rhetorical moves ensures that the generated analytical descriptions are structured to be a coherent analytical argument, and the sampling strategy encourages diversity in sentence structures.  
    \item Once the template sentences are selected, we replace the variables, entities, and values in the sentences with the given generated chart's meta-information. For example, consider the template sentence ``\textit{The [y-axis\_label] of [x-axis\_label] is observed to decline since [x-tick\_label]}.'', we substitute the variables with the generated chart's meta-information and generate the sentence ``\textit{The number of visitors of Singapore is observed to decline since 2015.}''. The script also analyzes corresponding relationships between data before performing the replacement if there is no related information in meta-data (i.e. the trend, statistical features such as minimum and maximum x-values and y-values, etc.). Such a process chooses templates randomly, and we can repeat the script three times to get multiple analytical chart descriptions for each chart. 
\end{enumerate}

Finally, the generated analytical descriptions are paired with the generated charts to form the \textsf{AutoChart} dataset.

\section{Dataset Evaluation}
\label{sec:empirical}
In order to conduct a thorough evaluation on the generated analytical descriptions, similar to many NLG tasks, we assess the generated analytical descriptions using both human and automatic metrics \cite{gatt2018survey}. 


\begin{figure}[t]
\centering

\begin{tabular}{p{18em}}
\multicolumn{1}{c}{\includegraphics[scale=0.18]{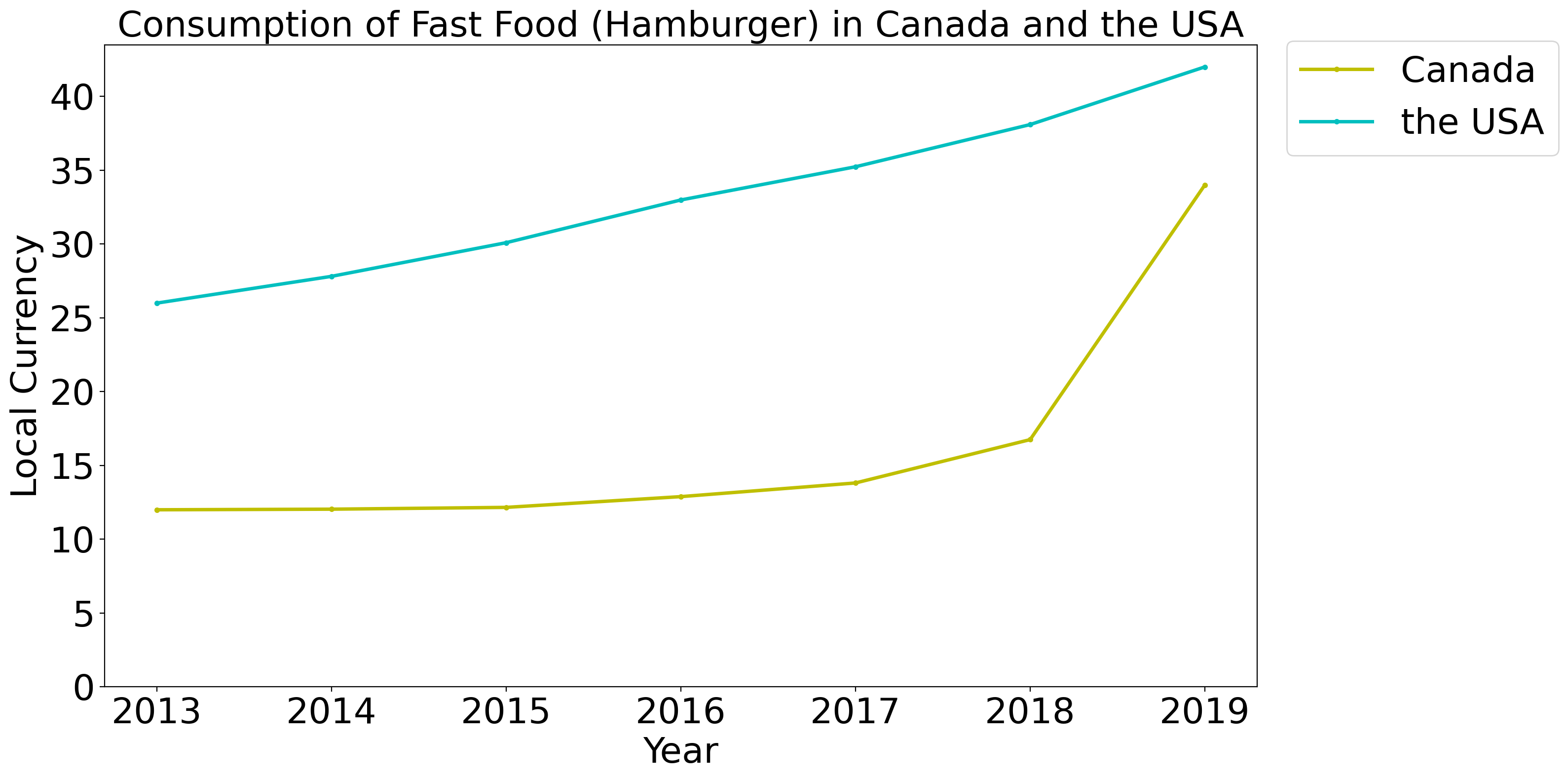}} \\ 

\textbf{Human:} \textit{\cellcolor[HTML]{EFEFEF} \small From 2013 to 2019, the line graph depicts the number of fast food (hamburger) consumption in Canada and the United States, respectively. In the last seven years, both countries have seen similar increases in consumer numbers. Over the last seven years, the United States has seen a steady increase. In 2018, there was a significant growth in Canada. Based on historical trends, both countries are anticipated to expand their fast food consumption in the coming years.}\\
\textbf{Generated:} \textit{\cellcolor[HTML]{E6FFDE} \small \textbf{[Move 1]} The line graph displays the number of consumption of fast food (hamburger) in Canada and the USA, respectively, from 2013 through 2019. \textbf{[Move 2]} In this chart, the unit of measurement is Local Currency, as seen on the y-axis. \textbf{[Move 3]} It is obvious that both countries shared similar increasing trends in the number of consumption in the past 6 years. \textbf{[Move 3.1]} For Canada, by 2013 the number of consumption reached nearly 12, while the number continued to increase until 34 in 2019. \textbf{[Move 3.1]} And for the USA, in 2013, the number of consumption was about 26, after that, each year has witnessed some increase.  \textbf{[Move 3]} In the past 6 years, the USA had consistently more than Canada. \textbf{[Move 5]} It would be interesting to see what would happen in the next decade in these two countries in terms of current situations.} \\

\end{tabular}\\
\caption{Example of a generated chart and the corresponding human and automatic generated analytical descriptions in \textsf{AutoChart} dataset.}
\label{fig:generated}
\end{figure}

\subsection{Dataset Overview}
Table \ref{tbl:pltbreakdown} summarizes our constructed \textsf{AutoChart} dataset. In total, we generated 10,232 charts and 23,543 corresponding analytical descriptions. Note that we have generated multiple analytical descriptions for each generated chart, simulating the real-world situation where different human writers may have different analytic descriptions of the same chart. The 150 analytical descriptions written by the linguistics researchers are also included in the dataset. The analytical descriptions have an average of 8 sentences and 140 words. Figure~\ref{fig:generated} shows an example of a generated chart and analytical description in the \textsf{AutoChart} dataset.





\subsection{Human Evaluation}
To examine the quality of the generated descriptions in \textsf{AutoChart}, we conducted three human-based evaluation studies. In the first study (S1), we recruited 30 linguistics researchers to write descriptions for 60 charts (20 line charts, 20 bar charts, and 20 scatter plots). The written descriptions from S1 are used in automatic evaluation discussed in the next section and also as the charts in S3. In Study 2 (S2) and Study 3 (S3), we examined the differences between \textsf{AutoChart} generated descriptions and the human-written descriptions from S1, respectively. They are the same otherwise in format and content. We studied S2 and S3 with 600 unique participants (20 line charts, 20 bar charts 20 scatter plots, each evaluated five times = 300 participants $\times$ 2 studies) using crowdsourcing on Amazon Mechanical Turk (AMT). Participants were at least 21 years old and were self-reported to be proficient in English. To reduce the potential bias in self-report, we used AMT's options to select only US-based workers. 

Informed consent was first obtained from participants. They then completed a demographics survey before proceeding to the study task. Participants were presented with a chart and its accompanying description, and then asked to rate the description on three dimensions of \textit{naturalness}, \textit{informativeness}, and \textit{quality} (i.e., grammatical correctness) adapted from the study in \cite{novikova2018rankme} using a 5-pt Likert scale. To ensure that participants were focused during the task, we asked them to answer a question that pertained to the chart description. We additionally used a reCAPTCHA~\cite{recaptcha} to reduce the likelihood of bot responses. Five participants rate each chart, and we compute the median to provide majority voting in ratings.

\begin{figure}[t]
    \centering
		\includegraphics[scale=0.8]{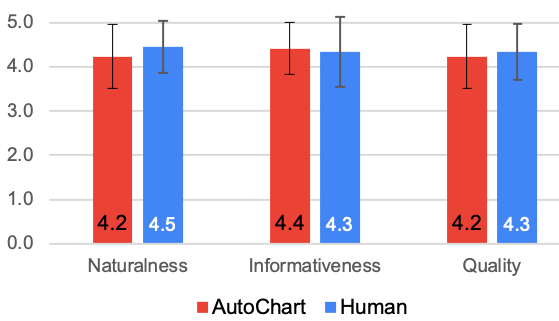}
	\caption{AutoChart vs Human descriptions rated on \textit{naturalness}, \textit{informativeness}, and  \textit{quality}}
	\label{fig:human-overall}
\end{figure}

\textbf{Results.} Comparing the results of S2 and S3, we did not detect significant differences between \textsf{AutoChart} and human-written descriptions for naturalness ($p = 0.056 > 0.05$, 1-tail), informativeness ($p = 0.288$) or quality ($p = 0.227$). From Figure~\ref{fig:human-overall}, we observe that human descriptions are rated higher on dimensions of naturalness and marginally on quality; with the generated analytical description in \textsf{AutoChart} performing marginally better on informativeness. No significant differences were also detected when the S2 and S3 were analysed at the chart type level. However, \textsf{AutoChart} had marginally better absolute performance on all three dimensions for \textit{bar} charts (respectively as (naturalness, informativeness, quality); \textsf{AutoChart}: (4.5, 4.6, 4.5) vs Human: (4.4, 4.4, 4.4)). \textsf{AutoChart} also performed marginally better on absolute informativeness for \textit{line} charts (4.6 vs 4.4). The results of the human-based evaluation suggest that the \textsf{AutoChart}'s generated analytical descriptions are similar to human-written descriptions in terms of informativeness, naturalness, and quality. 

\subsection{Automatic Evaluation}
Automatic evaluation of NLG tasks is challenging and an ongoing research area itself. The challenges of evaluating charts' analytical description automatically are compounded as the generated text are significantly longer than other NLG task such as machine translation. Nevertheless, we leverage existing automatic evaluation metrics commonly used in NLG tasks to evaluate our generated text. Specifically, we perform two automatic assessments on the \textsf{AutoChart} dataset: (i) Quality assessment, which compares the automatic generated analytical descriptions and 60 human-written references written by the linguistics researchers in human evaluation study S1. (ii) Difficulty assessment, where to train existing chart-to-text methods using the \textsf{AutoChart} dataset and compare their generated descriptions against the human-written references.


\subsection{Quality Assessment}
To evaluate the quality of the analytical descriptions in \textsf{AutoChart}, we computed the ROUGE~\cite{papineni2002bleu}, BLEU~\cite{lin2004rouge} and BLEURT~\cite{sellam2020bleurt} scores between the human-written references from earlier human-based evaluation study S1 and the automatic generated analytical descriptions for the same 60 charts. We assume that the human-written references are the gold standard, and the generated analytical descriptions in \textsf{AutoChart} should be similar to the gold standard.

\begin{table}[t]
\small
\centering
\begin{tabular}{l|c|c|c}
\hline
\textbf{Method} & \textbf{BLEU} & \textbf{ROUGE} & \textbf{BLEURT} \\ \hline \hline
\textbf{AutoChart} &  &  &  \\  
- Bar & 40.21 & 42.99 & 21.42 \\ 
- Line & 43.93 & 47.32 & 22.58 \\ 
- Scatter & 39.69 & 48.03 & 17.30 \\ 
- Overall & 41.28 & 46.11 & 20.43 \\\hline
\textbf{Baseline} &  &  &  \\  
- Bar & 32.63 & 35.95 & 12.25 \\ 
- Line & 35.48 & 33.20 & 7.55 \\ 
- Scatter & 32.28 & 32.33 & 9.12 \\ 
- Overall & 33.46 & 33.83 & 9.64 \\\hline
\end{tabular}
\caption{Quality Assessment Results.}
\label{tbl:quality_ass}
\end{table}

As a baseline comparison, we adopt a simple template-based generative method that generates the charts' analytical descriptions by randomly sampling the sentences from our templates and applying the charts' meta-data to produce the relevant analytical description. The main difference between the baseline and the \textsf{AutoChart} analytical descriptions is the baseline does not consider the rhetorical moves in the description generation. 

Table~\ref{tbl:quality_ass} shows the results of quality assessment on the analytical descriptions in \textsf{AutoChart} dataset and baseline. We compute the average scores for various automatic assessment metrics for the different chart types. The overall average scores are also reported. We observe that the \textsf{AutoChart}'s analytical descriptions significantly outperformed the baseline generated text, suggesting that the inclusion of rhetorical moves in analytical descriptions are more aligned to the human-written references.

\subsection{Difficulty Assessment}
Besides evaluating the quality of the \textsf{AutoChart} dataset, we are also interested in investigating the existing chart-to-text methods' performance in our new dataset. The goal is to assess the difficulty of generation chart analytical descriptions using the existing methods and the \textsf{AutoChart} dataset. Specifically, for this experiment, we first train the two state-of-the-art chart-to-text baselines~\cite{balaji2018chart, obeid2020chart} and an image captioning method~\cite{liu2020image} using the \textsf{AutoChart} dataset. Subsequently, we apply the trained baselines to generate the descriptions for the 60 charts in human evaluation study S1. Finally, we compute the ROUGE, BLEU, and BLEURT scores between the human-written references and the baselines' generated descriptions of the charts.

\begin{table}[t]
\small
\centering
\begin{tabular}{p{8em}|c|c|c}
\hline
\textbf{Method} & \textbf{BLEU} & \textbf{ROUGE} &  \textbf{BLEURT} \\ \hline \hline 
\citet{balaji2018chart} & 20.45 &22.9 & 13.31  \\\hline
\citet{obeid2020chart} & 33.05 & 28.32 & 18.23 \\\hline
\citet{liu2020image} & 10.68  & 19.74 & 5.49\\\hline
\end{tabular}
\caption{Difficulty Assessment Results.}
\label{tbl:diff_ass}
\end{table}

Table~\ref{tbl:diff_ass} shows the experiment results. We observe that none of the methods can perform exceeding well in generating chart descriptions that are close to human references. The best performing baselines, \cite{obeid2020chart}, was able to achieve similar results to the simple template-based generative baseline used in the quality assessment experiment. Unsurprisingly, the \cite{obeid2020chart} is not able to perform well for the chart analytical description generation task as the model did not consider the paragraph structure (i.e., rhetorical moves) in its generation. \cite{balaji2018chart} is designed to generate simple single sentence summaries for charts. Thus, it might not be able to generate informative and detailed analytical descriptions of the charts. The image caption method \cite{liu2020image} performed badly for the task as it is likely to generate the general captions such as ``\textit{this is a line chart.}''. The performance of existing baselines highlights the difficulty of the chart analytical description generation task.

\section{Discussion and Conclusion}
\label{sec:conclusion}
The \textsf{AutoChart} dataset opens up new research opportunities for the computer vision, computational linguistics, and natural language processing research communities. Novel object recognition and deep text generative models can be designed to interpret charts and generate relevant analytical descriptions automatically. The automatic interpretation and generation of analytical chart descriptions have many academic and industrial applications. For instance, generating good-quality analytic chart descriptions can guide students to attempt the IELTS AWT1. The automated analysis of charts is also a valuable function in existing assisted writing tools. The \textsf{AutoChart} dataset can support the development and exploration of the supervised chart-to-text methods.

We opined that this is the start of an emerging research topic, and many future works could be done. As an extension of this work, we aim to investigate and model more sophisticated linguistic techniques to construct better quality analytical descriptions of charts. We will expand the dataset to include more types of charts, e.g., pie charts, box plots, etc. Finally, we will also explore more automatic evaluation methods to assess the quality of the generated analytical descriptions. For example, we can examine and assess the analytical descriptions' logic, reasoning, and fluency.

To conclude, we have proposed a novel framework that automatically constructs the \textsf{AutoChart} dataset, a large chart analytical description dataset.  We conducted extensive human and machine evaluation on the generated charts and descriptions and demonstrate that the generated text is informative, coherent and relevant to the corresponding charts. We hope that the \textsf{AutoChart} can encourage more research in the automatic generation of analytical descriptions of charts.

\section*{Acknowledgement}
This research is supported by Living Sky Technologies Ltd, Canada under its research exploratory funding initiatives. Any opinions, findings and conclusions or recommendations expressed in this material are those of the author(s) and do not reflect the views of Living Sky Technologies Ltd, Canada.


\bibliographystyle{acl_natbib}
\bibliography{ranlp2021}

\begin{thebibliography}{24}
\expandafter\ifx\csname natexlab\endcsname\relax\def\natexlab#1{#1}\fi

\bibitem[{rec()}]{recaptcha}

\newblock {reCAPTCHA}.
\newblock \url{https://www.google.com/recaptcha/about}.
\newblock Accessed: 2020-11-21.

\bibitem[{Balaji et~al.(2018)Balaji, Ramanathan, and Sonathi}]{balaji2018chart}
Abhijit Balaji, Thuvaarakkesh Ramanathan, and Venkateshwarlu Sonathi. 2018.
\newblock Chart-text: A fully automated chart image descriptor.
\newblock \emph{arXiv preprint arXiv:1812.10636}.

\bibitem[{Battle et~al.(2018)Battle, Duan, Miranda, Mukusheva, Chang, and
  Stonebraker}]{battle2018beagle}
Leilani Battle, Peitong Duan, Zachery Miranda, Dana Mukusheva, Remco Chang, and
  Michael Stonebraker. 2018.
\newblock Beagle: Automated extraction and interpretation of visualizations
  from the web.
\newblock In \emph{Proceedings of the 2018 CHI Conference on Human Factors in
  Computing Systems}, pages 1--8.

\bibitem[{Chai et~al.(2020)Chai, Li, Fan, and Luo}]{chai2020crowdchart}
Chengliang Chai, Guoliang Li, Ju~Fan, and Yuyu Luo. 2020.
\newblock Crowdchart: Crowdsourced data extraction from visualization charts.
\newblock \emph{IEEE Transactions on Knowledge and Data Engineering}.

\bibitem[{Cliche et~al.(2017)Cliche, Rosenberg, Madeka, and
  Yee}]{cliche2017scatteract}
Mathieu Cliche, David Rosenberg, Dhruv Madeka, and Connie Yee. 2017.
\newblock Scatteract: Automated extraction of data from scatter plots.
\newblock In \emph{Joint European Conference on Machine Learning and Knowledge
  Discovery in Databases}, pages 135--150. Springer.

\bibitem[{Dai et~al.(2018)Dai, Wang, Niu, and Zhang}]{dai2018chart}
Wenjing Dai, Meng Wang, Zhibin Niu, and Jiawan Zhang. 2018.
\newblock Chart decoder: Generating textual and numeric information from chart
  images automatically.
\newblock \emph{Journal of Visual Languages \& Computing}, 48:101--109.

\bibitem[{Gatt and Krahmer(2018)}]{gatt2018survey}
Albert Gatt and Emiel Krahmer. 2018.
\newblock Survey of the state of the art in natural language generation: Core
  tasks, applications and evaluation.
\newblock \emph{Journal of Artificial Intelligence Research}, 61:65--170.

\bibitem[{Hunter(2007)}]{matplotlib}
J.D. Hunter. 2007.
\newblock Matplotlib: A 2d graphics environment.

\bibitem[{Iso et~al.(2019)Iso, Uehara, Ishigaki, Noji, Aramaki, Kobayashi,
  Miyao, Okazaki, and Takamura}]{Iso2019Learning}
Hayate Iso, Yui Uehara, Tatsuya Ishigaki, Hiroshi Noji, Eiji Aramaki, Ichiro
  Kobayashi, Yusuke Miyao, Naoaki Okazaki, and Hiroya Takamura. 2019.
\newblock Learning to select, track, and generate for data-to-text.
\newblock In \emph{Proceedings of the 57th Annual Meeting of the Association
  for Computational Linguistics (ACL)}.

\bibitem[{Kahou et~al.(2017)Kahou, Michalski, Atkinson, K{\'a}d{\'a}r,
  Trischler, and Bengio}]{kahou2017figureqa}
Samira~Ebrahimi Kahou, Vincent Michalski, Adam Atkinson, {\'A}kos
  K{\'a}d{\'a}r, Adam Trischler, and Yoshua Bengio. 2017.
\newblock Figureqa: An annotated figure dataset for visual reasoning.
\newblock \emph{arXiv preprint arXiv:1710.07300}.

\bibitem[{Karatzas~I.(1998)}]{brownianmotion}
Shreve~S.E. Karatzas~I. 1998.
\newblock Brownian motion. in: Brownian motion and stochastic calculus.
\newblock \emph{Graduate Texts in Mathematics}.

\bibitem[{Lepp{\"a}nen et~al.(2017)Lepp{\"a}nen, Munezero, Granroth-Wilding,
  and Toivonen}]{leppanen2017data}
Leo Lepp{\"a}nen, Myriam Munezero, Mark Granroth-Wilding, and Hannu Toivonen.
  2017.
\newblock Data-driven news generation for automated journalism.
\newblock In \emph{Proceedings of the 10th International Conference on Natural
  Language Generation}, pages 188--197.

\bibitem[{Lin(2004)}]{lin2004rouge}
Chin-Yew Lin. 2004.
\newblock Rouge: A package for automatic evaluation of summaries.
\newblock In \emph{Text summarization branches out}, pages 74--81.

\bibitem[{Liu et~al.(2020)Liu, Li, Hu, Guan, and Tian}]{liu2020image}
Maofu Liu, Lingjun Li, Huijun Hu, Weili Guan, and Jing Tian. 2020.
\newblock Image caption generation with dual attention mechanism.
\newblock \emph{Information Processing \& Management}, 57(2):102178.

\bibitem[{Liu et~al.(2019)Liu, Klabjan, and NBless}]{liu2019data}
Xiaoyi Liu, Diego Klabjan, and Patrick NBless. 2019.
\newblock Data extraction from charts via single deep neural network.
\newblock \emph{arXiv preprint arXiv:1906.11906}.

\bibitem[{Ma et~al.(2018)Ma, Tung, Wang, Gao, Pan, and Chen}]{ma2018scatternet}
Yuxin Ma, Anthony~KH Tung, Wei Wang, Xiang Gao, Zhigeng Pan, and Wei Chen.
  2018.
\newblock Scatternet: A deep subjective similarity model for visual analysis of
  scatterplots.
\newblock \emph{IEEE transactions on visualization and computer graphics}.

\bibitem[{Methani et~al.(2020)Methani, Ganguly, Khapra, and
  Kumar}]{methani2020plotqa}
Nitesh Methani, Pritha Ganguly, Mitesh~M Khapra, and Pratyush Kumar. 2020.
\newblock Plotqa: Reasoning over scientific plots.
\newblock In \emph{The IEEE Winter Conference on Applications of Computer
  Vision}, pages 1527--1536.

\bibitem[{Molle and Prior(2008)}]{molle2008multimodal}
Daniella Molle and Paul Prior. 2008.
\newblock Multimodal genre systems in eap writing pedagogy: Reflecting on a
  needs analysis.
\newblock \emph{Tesol Quarterly}, 42(4):541--566.

\bibitem[{Novikova et~al.(2018)Novikova, Du{\v{s}}ek, and
  Rieser}]{novikova2018rankme}
Jekaterina Novikova, Ond{\v{r}}ej Du{\v{s}}ek, and Verena Rieser. 2018.
\newblock Rankme: Reliable human ratings for natural language generation.
\newblock \emph{arXiv preprint arXiv:1803.05928}.

\bibitem[{Obeid(2020)}]{obeid2020chart}
Jason Obeid. 2020.
\newblock Chart-to-text: Generating natural language descriptions for charts by
  adapting the transformer model.
\newblock In \emph{Proceedings of the 13th International Conference on Natural
  Language Generation}, pages 138--147.

\bibitem[{Papineni et~al.(2002)Papineni, Roukos, Ward, and
  Zhu}]{papineni2002bleu}
Kishore Papineni, Salim Roukos, Todd Ward, and Wei-Jing Zhu. 2002.
\newblock Bleu: a method for automatic evaluation of machine translation.
\newblock In \emph{Proceedings of the 40th annual meeting of the Association
  for Computational Linguistics}, pages 311--318.

\bibitem[{Puduppully et~al.(2019)Puduppully, Dong, and
  Lapata}]{DBLP:journals/corr/abs-1809-00582}
Ratish Puduppully, Li~Dong, and Mirella Lapata. 2019.
\newblock \href {http://arxiv.org/abs/1809.00582} {Data-to-text generation with
  content selection and planning}.
\newblock In \emph{Proceedings of the 33rd AAAI Conference on Artificial
  Intelligence}, Honolulu, Hawaii.

\bibitem[{Sellam et~al.(2020)Sellam, Das, and Parikh}]{sellam2020bleurt}
Thibault Sellam, Dipanjan Das, and Ankur~P Parikh. 2020.
\newblock Bleurt: Learning robust metrics for text generation.
\newblock \emph{arXiv preprint arXiv:2004.04696}.

\bibitem[{Swales(2004)}]{swales2004research}
John~M Swales. 2004.
\newblock \emph{Research genres: Explorations and applications}.
\newblock Cambridge University Press.

\end{thebibliography}


\end{document}